%% file: main.tex
\documentclass[conference]{IEEEtran}
\IEEEoverridecommandlockouts
\usepackage{cite}
\usepackage{amsmath,amssymb,amsfonts}
\usepackage{algorithmic}
\usepackage{graphicx}
\usepackage{textcomp}
\usepackage{xcolor}

\usepackage{multirow}
\usepackage[hang, tight]{subfigure}
\usepackage{wrapfig}
\usepackage{hyperref}
\usepackage{url}
\usepackage{comment}
\usepackage{color}
\usepackage{etoolbox}
\usepackage[ruled,vlined]{algorithm2e}
\usepackage{amssymb}
\usepackage{bm}

\def\BibTeX{{\rm B\kern-.05em{\sc i\kern-.025em b}\kern-.08em
    T\kern-.1667em\lower.7ex\hbox{E}\kern-.125emX}}
\begin{document}

\title{
DASNet: Dynamic Activation Sparsity for Neural Network Efficiency Improvement
}

\author{\IEEEauthorblockN{Qing Yang$^1$, Jiachen Mao$^1$, Zuoguan Wang$^2$ and Hai Li$^1$}
\IEEEauthorblockA{
\textit{$^1$Department of Electrical and Computer Engineering, Duke University, Durham, North Carolina, USA} \\
\textit{$^2$Black Sesame Technologies, Santa Clara, California, USA} \\
$^1$\{qing.yang21, jiachen.mao, hai.li\}@duke.edu, $^2$zuoguan.wang@bst.ai
}}

\maketitle

\begin{abstract}
To improve the execution speed and efficiency of neural networks in embedded systems, it is crucial to decrease the model size and computational complexity.
In addition to conventional compression techniques, e.g., weight pruning and quantization, removing unimportant activations can reduce the amount of data communication and the computation cost. 
Unlike weight parameters, the pattern of activations is directly related to input data and thereby changes dynamically. 
To regulate the dynamic activation sparsity (DAS), in this work, we propose a generic low-cost approach based on winners-take-all (WTA) dropout technique. 
The network enhanced by the proposed WTA dropout, namely \textit{DASNet}, features structured activation sparsity with an improved sparsity level. 
Compared to the static feature map pruning methods, DASNets provide better computation cost reduction. 
The WTA technique can be easily applied in deep neural networks without incurring additional training variables. 
More importantly, DASNet can be seamlessly integrated with other compression techniques, such as weight pruning and quantization, without compromising on accuracy. 
Our experiments on various networks and datasets present significant run-time speedups with negligible accuracy loss. 
\end{abstract}


\input{intro.tex}
\input{related_works.tex}
\input{approach.tex}
\input{evaluation.tex}
\input{conclusion.tex}

\bibliographystyle{./IEEEtran}
\bibliography{./IEEEexample}

\end{document}

%% file: intro.tex
\section{Introduction}

With rapid performance improvement in cognitive tasks, \textit{deep neural networks} (DNNs) have become a main horsepower of \textit{artificial intelligence} (AI) applications.
In spite of the promising success on large-scale CPU- and GPGPU-based platforms, the embedded AI is facing rigorous challenges. 
First, the limited on-chip memory can accommodate only a small portion of model parameters and inter-layer data, inducing continuous data exchanges with the off-chip main memory. 
Moreover, the intensive 2-D convolutions demand high computational complexity and inter-layer data traffic. 
For example, to process one image, AlexNet~\cite{krizhevsky2012imagenet} owning a model size of 243MB involves about 748M \textit{multiply-and-accumulate} (MAC) operations. 
To overcome the challenges in large model size and high computational complexity of DNNs, \textit{weight quantization} and \textit{pruning} have been widely utilized in embedded AI deployment. 

The quantization is a popular approach to reduce model size by lowering the weight precision (aka bit-width), \textit{e.g.}, to 8-bit integer~\cite{jouppi2017datacenter} or even binary levels~\cite{courbariaux2015binaryconnect}. 
BinaryNet~\cite{hubara2016binarized} and XNOR net~\cite{rastegari2016xnor} further applied the quantization to neuron activations to reduce inter-layer data traffic. 
Low-precision operations are more efficient when implementing on hardware.
For example, the power consumption of 8-bit integer multiplication is only 5.4\% of the 32-bit floating-point counterpart~\cite{horowitz20141}. 
Inevitably, deeply quantized models suffer from performance loss.
For example, more than $12\%$ accuracy drop was observed when applying XNOR net on ImageNet~\cite{rastegari2016xnor}.

Weight pruning is usually realized by applying $\ell1$ or $\ell2$ regularization on the weights. 
Wen \textit{et al.} \cite{wen2016learning} proposed \textit{structured sparsity learning} (SSL) which utilizes group lasso to get rid of redundant weight groups with user-defined shapes in convolution layers. 
Molchanov \textit{et al.}~\cite{molchanov2017pruning} included the first-order Taylor series expansion of the loss function on feature map channels to determine unimportant channels before filter pruning. 
It's worth mentioning that the weight pruning degrades the model's representation capability and thereby hinders the utilization of low-precision weights. 
As an example of integrating model quantization and weight pruning,
Han \textit{et al.}~\cite{han2015deep} managed to fine-tune a dedicated non-linear code book for each layer to indicate the weight sharing. 
The quantized model still relies on the weights in a high-precision level (\textit{e.g.}, 32-bit fixed-point) after looking up the actual weights from nonlinear code books. 

Activation sparsity is another essential feature that can be used for DNN acceleration. 
The widely adopted activation function, \textit{rectified linear unit} (ReLU), can be taken as a formation of activation sparsity, as it produces many zero activations.
It brings the potential of reducing the inter-layer data movement by integrating with data compression. 
For example, Eyeriss~\cite{chen2017eyeriss} implemented an on-chip run-length compression module, which achieved up to 1.9$\times$ reduction in memory accesses. 
Rhu \textit{et al.}~\cite{rhu2018compressing} proposed a dedicated \textit{compressing direct memory access} (cDMA) engine to exploit the inherent sparsity in activations to speedup the data movements between the CPU memory and the GPU memory. 
In addition, EIE~\cite{han2016eie} and ZeNA~\cite{kim2018zena} utilized non-zero detection logic unit to select non-zero activations for processing engines, achieving significant energy saving and speedup.

\begin{figure}[t]
\begin{minipage}{\columnwidth}
    \centering
    \subfigure[Original.]{
    \includegraphics[width=0.4\columnwidth]{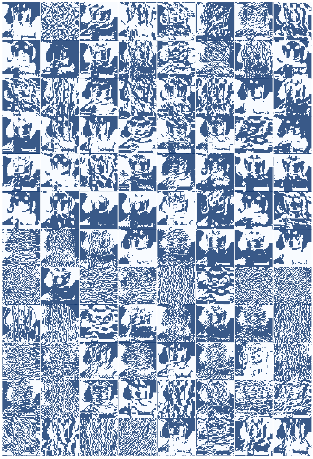}}
    \hspace{5pt}
    \subfigure[With WTA masks.]{
    \includegraphics[width=0.4\columnwidth]{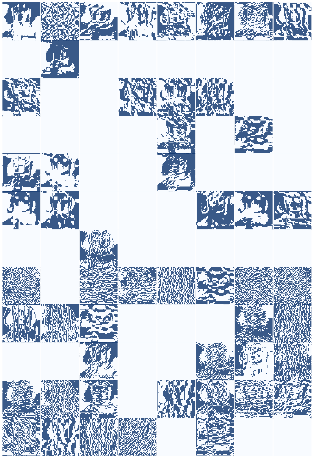}}
    \vspace{-6pt}
    \caption{An example of the activation distribution of the first convolution layer in AlexNet on ImageNet dataset.}
    \vspace{-6pt}
    \label{fig:fm}
\end{minipage}
\end{figure}

Although the activation sparsity has great potential in saving computation cost and power consumption, two fundamental issues prevent the utilization of its full advantage in practice.
\textit{(i)} The intrinsic activation sparsity pattern from ReLU is random. 
Fig.~\ref{fig:fm}(a) is an example of the flattened feature map after the first convolution layer in AlexNet on ImageNet dataset, where the white dots stand for zero activations.
To leverage the random activation sparsity distribution, specific hardware designs are compulsive. 
In other words, it is hard to achieve the equivalent speedup on conventional embedded systems without the support of dedicated circuit components. 
\textit{(ii)} The sparsity levels for different input images are not identical. 
The overhead of non-zero detection and compression/decompression techniques can be amortized for sparse activations, which might not be true for dense activations incurred by certain input data. 


This work aims to overcome the aforementioned difficulties, enable the feasible activation sparsification and improve the efficiency of DNN deployment in embedded systems.
We exploit the \textit{dynamic activation sparsity} (DAS) and propose a \textit{dynamic winners-take-all (WTA) dropout} technique. 
In each layer, the neurons will be tagged with rankings based on their activation magnitudes at run-time. 
Only those neurons with top rankings will be selected to participate in the computations for the following layers. 
In other words, the WTA dropout serves as a dynamic WTA mask to prune neuron activations layer-wise. 
Thus, we are able to obtain the structured activation sparsity by removing certain channels in feature maps for improving the sparsity level, as shown in Fig.~\ref{fig:fm}(b). 
Moreover, the rate of WTA dropout can be adjusted layer by layer, ensuring the overall system accuracy.
Our experiments on various datasets show that the derived networks enhanced by WTA dropout, namely \textit{DASNets}, can achieve $1.12\times\sim1.8\times$ speedup by paying at most $0.5\%$ accuracy drop.
In addition, the DASNets can be integrated with state-of-the-art model compression techniques, such as weight pruning and quantization, to further improve the execution efficiency.
For example, combined with weight pruning (or 8-bit quantization), the DASNet obtained based on AlexNet can be safely compressed $4.7\times$ (or $4\times$) without compromising on the accuracy. 


%% file: related_works.tex
\section{Related Works}
\label{sec:Related}

Some DNN accelerator architectures manage to regulate the neuron activation sparsity beyond the intrinsic sparse patterns obtained from ReLU.
For example, Cnvlutin~\cite{albericio2016cnvlutin} and Minerva~\cite{reagen2016minerva} removed activations with small magnitudes to further increase the activation sparsity level. 
Without model accuracy loss, however, the increment in the activation sparsity is limited. 
Moreover, dedicated circuit modules for the data compression and decompression are required on irregular sparse patterns, which shares the similarity with Eyeriss~\cite{chen2017eyeriss} and cDMA engine~\cite{rhu2018compressing}.

At the algorithmic level, Spring \textit{et al.}~\cite{spring2017scalable} utilized the \textit{locality-sensitive hashing} (LSH) to predict the neurons with top activation magnitudes for DNNs with only fully-connected (fc) layers. 
In both training and inference stages, only the selected neurons are activated.
The LSH-based method produces a higher sparsity in activations (\textit{e.g.}, on average only 5\% activations remain).
However, it is hard to be applied to convolution (conv) layers.
%
Feature map pruning has been studied. 
For example, Molchanov \textit{et al.}~\cite{molchanov2017pruning} analyzed the importance of feature map channels based on the partial derivative of the loss function and eliminated those less important ones. 
The scaling factor in batch normalization can also be utilized to indicate the significance of each feature map channel~\cite{liu2017learning,ye2018rethinking}. 
The idea of adding scaling factors on feature maps can be extended to any conv layers even without batch normalization~\cite{huang2018data}. 
In the training stage, all scaling factors are learned by adding a specified regularization term in the original loss function. 
A feature map channel whose scaling factor is less than a predefined threshold can be safely removed once the training is completed. 
The partial selection of feature map channels can also be dynamically determined at run-time during inference to optimize the accuracy loss~\cite{hua2018channel,gao2018dynamic}. 
However, the existing dynamic approaches introduce intricate branching structures built with the original model, which increases the model size and complicates the training process. 

Compared to the existing techniques, our proposed DASNet comes with three major improvements: 
higher sparsity levels with structured sparse patterns, 
simpler training process without additional regularization terms, and no need of extra trainable parameters beyond the original models. 
The details of our method will be described in the following sections. 

%% file: approach.tex
\section{DASNet with WTA Dropout}
\label{sec:Approach}

\begin{figure}[t]
\begin{minipage}{\columnwidth}
    \centering
    \subfigure[Original.]{
    \includegraphics[width=0.48\columnwidth]{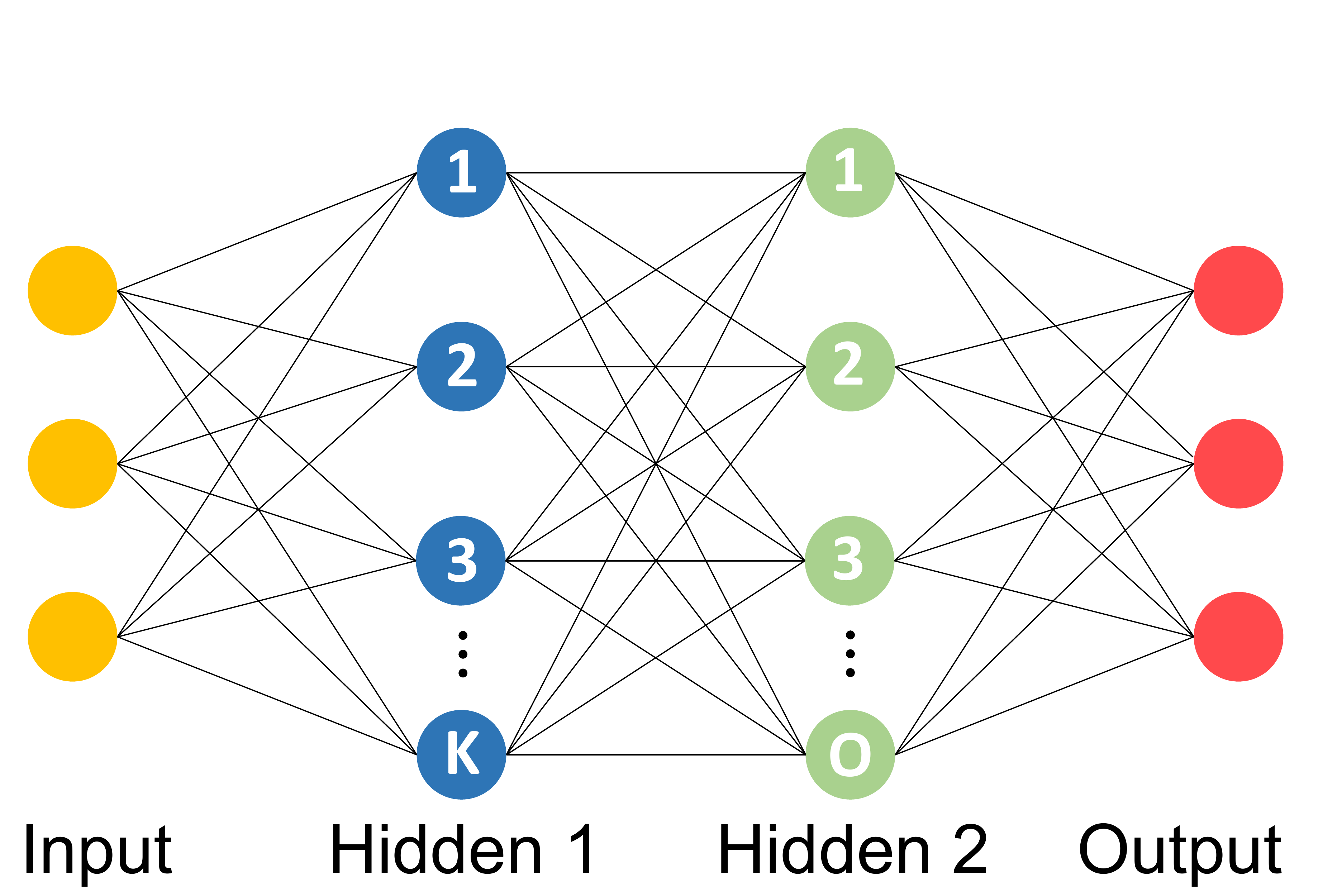}}
    \subfigure[With WTA masks.]{
    \includegraphics[width=0.48\columnwidth]{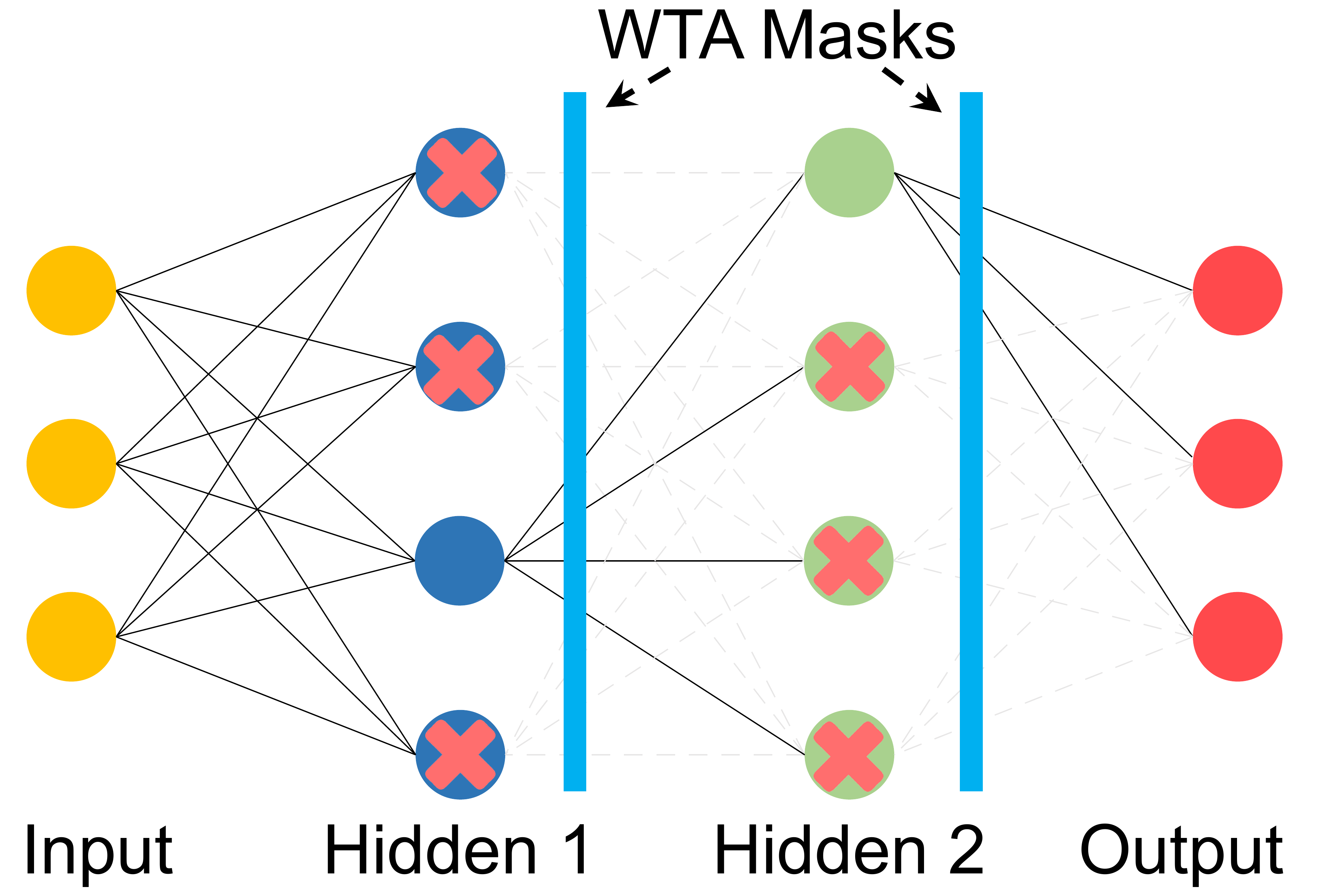}}
    \caption{The DASNet with dynamic WTA dropout.}
    \vspace{-9pt}
    \label{fig:fc_wta}
\end{minipage}
\end{figure}

In this work, we propose \textit{dynamic WTA dropout} which regulates the neuron activation sparsity based on the activation magnitudes, as
stronger activations potentially contribute more to the computation results in both forward and backward propagation phases. 
Fig. \ref{fig:fc_wta} depicts the basic concept of the proposed dynamic WTA dropout. 
It behaves as a mask between layers and prunes low-ranking neurons at run-time. 
We first rank all the neurons of a layer based on their activation magnitudes.
Only those top-ranking neurons (namely, \textit{winner neurons}) will remain and propagate their outputs to the following layer. 
DASNets with WTA masks need to be finetuned in order to maintain the model accuracy. 
In the backward propagation phase during finetuning, accordingly, we update only the winner neurons selected by the WTA masks along the forward path. 
%
The \textit{winner rate} ($p$) of a layer is defined as: 
\begin{equation}
\label{eq:define_p}
p = N_{winner}/N_{total}, 
\end{equation}
where $N_{winner}$ and $N_{total}$ denote the sizes of the winner neuron set $\mathrm{S_{winner}}$ and the total neuron set $\mathrm{S_{total}}$, respectively. 
$p$ is adjustable to balance the sparsity level and accuracy. 
The finetuning process is clarified in \textbf{Algorithm \ref{alg1}}, which shares the same loss function used for training the original model. 
Because of the WTA dropout, a new term $\frac{\partial A_{w, l}}{\partial A_{o, l}}$ will be applied on the weight updates during the backpropagation phase. Actually, $\frac{\partial A_{w, l}}{\partial A_{o, l}}$ is the binary mask generated by WTA dropout, determining which neurons should be updated for the particular input data $x_i$. 

\begin{algorithm}[b]
\small
\caption{DASNet Finetuning. \bm{$Loss$} is the loss function with input-output pair (\bm{$x_i, y_i$}), and the learning rate is \bm{$\eta$}. \bm{$W_l$} denotes the weight variables per layer. \bm{$A_{o,l}$} and \bm{$A_{w,l}$} are the original activation and pruned activation after the WTA mask, respectively. The total layer number is \bm{$L$}.}
\label{alg1}
\SetAlgoLined
\While{not reaching the stop criteria}
{
    \textbf{1. Forward Propagation:}\\
    For layer $l=1$, compute $A_{o, 1}=g(x_{i}, W_{1})$\;
    \For{$l=2\rightarrow L$}
        {get the activation $A_{w, l-1}$ for $S_{winner, l-1}$\; 
        compute $A_{o, l}=g(A_{w, l-1}, W_{l})$;}
    \textbf{2. Backward Propagation:}\\
    For layer $l=L$, $A_{w, L}=A_{o, L}$\;
    \For{$l=(L-1)\rightarrow 1$}
        {propagate $\frac{\partial Loss}{\partial A_{w, l}}$\;
        update $W_{l}=W_{l}+\Delta W_{l}$, where \\
        $\Delta W_{l}=-\eta \frac{\partial Loss}{\partial W_{l}}=-\eta \frac{\partial Loss}{\partial A_{w, l}}\cdot \frac{\partial A_{w, l}}{\partial A_{o, l}}\cdot \frac{\partial A_{o, l}}{\partial W_{l}}$\;}
}
\end{algorithm}

\subsection{Ranking and Winner Rate Selection}
\label{sec:winner_select}

The selection of the winner rate $p$ for each layer is a tradeoff process between the activation sparsification level and the model inference accuracy. 
From the one hand, a DASNet with more activations pruned potentially obtains faster acceleration; 
from the other hand, it becomes harder to keep the accuracy of such a DASNet intact.
To explore the relation between model accuracy and activation pruning strength (i.e., $p$), we first adopt three basic networks (MLP-3, LeNet-4, and ConvNet-5) on two simple datasets (MNIST and CIFAR-10).
More experiment for deep networks will be presented in Section~\ref{sec:Evaluation}.
Table \ref{tab:toy_models} summarizes the configurations of the three models. 
The thumb rule to determine winner neurons in conv and fc layers is also explained as follows. 

\begin{table}[t]
\centering
\caption{The network configuration.}
\label{tab:toy_models}
\resizebox{1\columnwidth}{!}
{\begin{minipage}{\columnwidth}
\centering
\begin{tabular}{cccc}
\hline\hline
Model & MLP-3  & LeNet-4    & ConvNet-5 \\
\hline
Dataset  & MNIST  & MNIST      & CIFAR-10  \\
Accuracy & 98.4\% &  99.4\%  &  86.0\%   \\  
\hline
conv1 & -   & 5$\times$5, 32   & 5$\times$5, 64  \\
conv2 & -   & 5$\times$5, 64   & 5$\times$5, 64  \\
\hline
fc1 & 784$\times$300    & 3136$\times$1024 & 2304$\times$384 \\
fc2 & 300$\times$100    & 1024$\times$10   & 384$\times$192  \\
fc3 & 100$\times$10     & -          & 192$\times$10 \\
\hline\hline
\end{tabular}
\end{minipage}}
\end{table}

\subsubsection{Fully-connected layers}

In fc layers, each neuron generates one single activation. 
The rankings of their activation magnitudes can be directly used to determine the winner neurons. 
The winner neuron selection is a relaxed \textit{partial sorting} problem. 
It can be solved by using recursive method such as~\cite{bentley1980general} in linear time complexity $\mathcal{O}(n)$, or $\mathcal{O}(n\log{}n)$ under the worst-case condition. 
Here, $n$ is equal to $N_{total}$ as defined in Equation~(\ref{eq:define_p}). 

The standard dropout~\cite{srivastava2014dropout} has been used to avoid overfitting for large fc layers. 
Our experiments is conducted by including the standard dropout with a dropout rate of 50\%.
We observe that although half of the neurons are activated in each layer, the proportion of neurons with non-zero activation is usually close to or even below 20\%. 
That is, only a small portion of neurons in a layer fire each time. 
Furthermore, the same dropout rate is usually applied across all the layers of a model, without considering the different sparsity levels presented by different layers.
Instead, our method determines $p$ in WTA masks according to the cumulative energy $E_{cum}$ of the winner set $\mathrm{S_{winner}}$, such as:
\begin{equation}
\label{eq:define_energy}
E_{cum}=\frac{\sum_{a_j\in \mathrm{S_{winner}}}{a_{j}}^{2}}{\sum_{a_i\in \mathrm{S_{total}}}{a_{i}}^{2}}, 
\end{equation}
where $a_i$ and $a_j$ denote the neuron activation. 
We set an threshold $\theta$. 
The size of $\mathrm{S_{winner}}$ and the winner rate $p$ then can be decided by satisfying the constraint of $E_{cum} \geq \theta$. 

\begin{figure}[t]
\centering
\includegraphics[width=0.8\columnwidth]{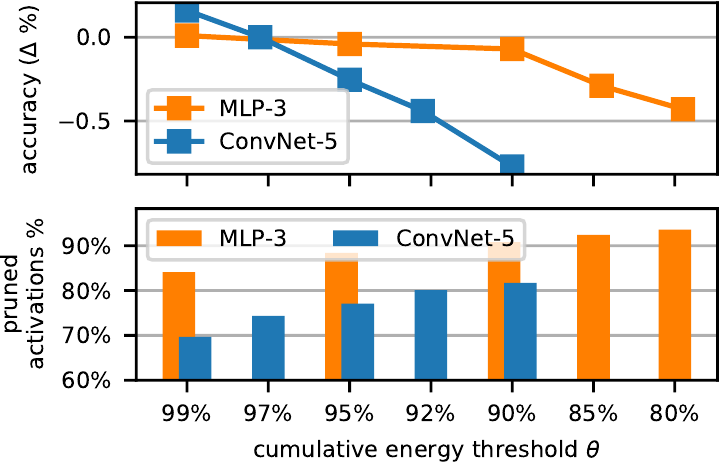}
\vspace{-6pt}
\caption{Accuracy vs. activation pruning for fc layers.}
\vspace{-12pt}
\label{fig:acc_fc}
\end{figure}

MLP-3 and ConvNet-5 are adopted to explore the appropriate energy threshold $\theta$ and the effects of winner rate $p$ on model accuracy. 
WTA masks are attached after all the fc layers except the output layer. 
The activation sparsity level is controlled by configuring $p$ according to the setting of $\theta$. 
Fig.~\ref{fig:acc_fc} shows the changes of model accuracy and the corresponding activation pruning strength when varying $\theta$. 
The pruned percentage of activations directly reflects the sparsity level and inter-layer data traffic reduction and is used to measure the pruning strength here. 
As expected, the model accuracy has an explicit inverse correlation with the pruning strength. 
Majority of the activations can be removed by scarifying a slight accuracy drop.
For instance, pruning 93.6\% activations from MLP-3 induces 0.4\% accuracy drop, and ConvNet-5 can mask out 81.7\% activations in its fc layers with 0.8\% accuracy drop.
It's worth noting that ConvNet-5 is more sensitive to activation pruning compared to MLP-3 due to the higher complexity of CIFAR-10 dataset. 
For this reason, different models require different constraints on $\theta$ in practice. 

\subsubsection{Convolution layers}
In conv layers, each filter is regarded as a neuron. 
Every neuron generates a channel in a feature map. 
We cannot directly adopt the ranking method used in fc layers as it is based on single activation. 
More specific, it is inefficient to rank all the activations in the feature map, as it involves with significant computation resources. 
We propose to get the winner set $\mathrm{S_{winner}}$ in a conv layer through a two-step ranking process: 
\textit{(i)} construct a feature vector ($fv$) with elements representing the importance of feature map channels. The vector length is equal to the number of filters; 
and \textit{(ii)} apply the ranking method used in fc layers to $fv$. 

\begin{figure}[t]
\begin{minipage}{\columnwidth}
\centering
\subfigure[Original.]{
\includegraphics[width=0.48\columnwidth]{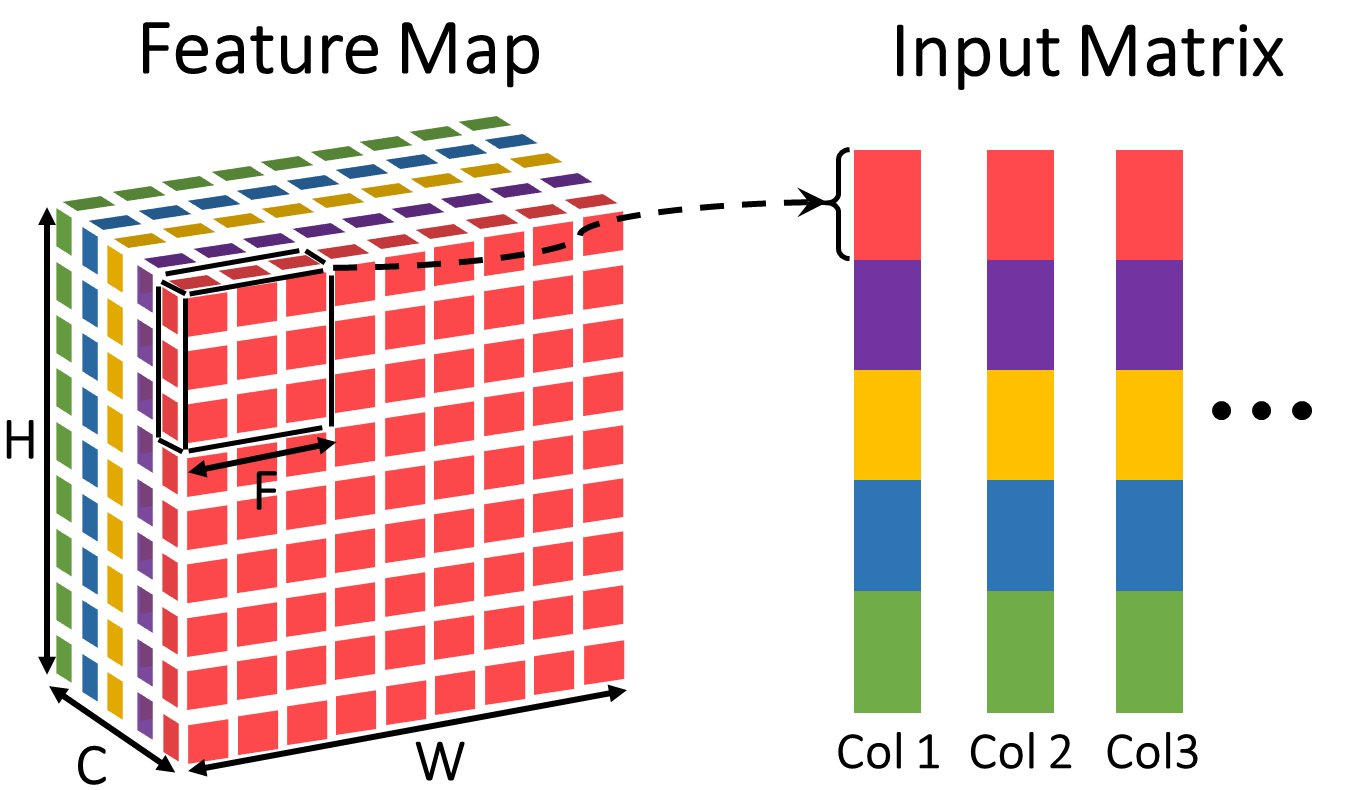}}
\subfigure[With WTA masks.]{
\includegraphics[width=0.48\columnwidth]{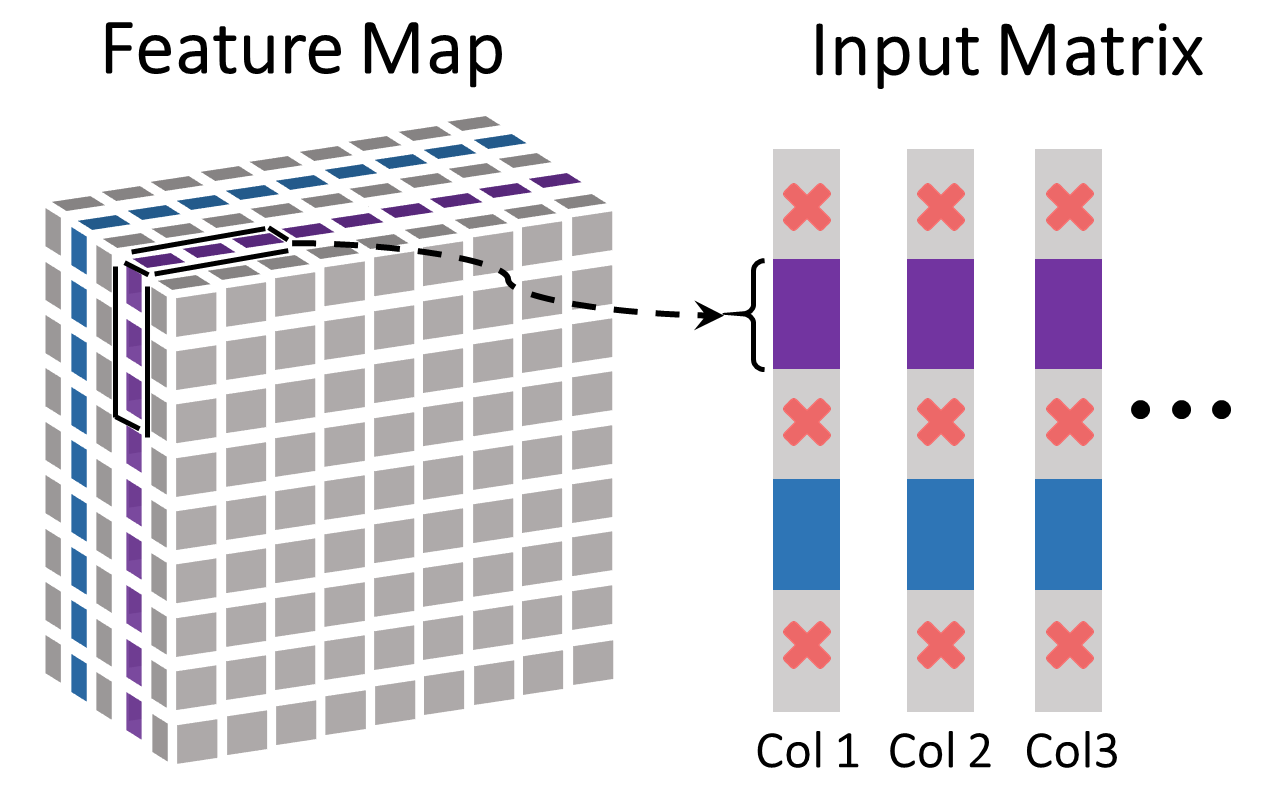}}
\caption{Working scheme of WTA dropout in conv layers. The feature map is a 3-D matrix with dimensions of height $H$, width $W$ and channel $C$.}
\vspace{-10pt}
\label{fig:conv_wta}
\end{minipage}
\end{figure}

There are two efficient ways to obtain the feature vector. 
Considering a feature map $fm\in \Re^{H\times W\times C}$ as depicted in Fig. \ref{fig:conv_wta}(a), the feature vector $fv=[v_1, v_2,\ldots, v_C] \in \Re^C$ can be constructed with the mean activation per channel: 
\begin{equation}
\label{eq:mean}
v_j = \mathbf{mean}(fm[:,:,j])\mbox{, where}\ j=1\ldots C.
\end{equation}
As previous studies~\cite{zeiler2014visualizing,bau2017network} discovered that the maximum activation in a feature map likely dominates the feature representation in the corresponding feature dimension. 
$fv$ can also be abstracted from $fm$ by: 
\begin{equation}
\label{eq:max}
v_j = \mathbf{max}(fm[:,:,j])\mbox{, where}\ j=1\ldots C.
\end{equation}

Each feature map channel can be regarded as a feature dimension.
Thus the eigenvalue for a feature dimension is a good indicator for the significance of a feature map channel. 
We propose to use the eigenvalues $\Lambda=\{\lambda_c, c=1\ldots C\}$ to replace $a_i$ and $a_j$ in Equation~(\ref{eq:define_energy}) to determine $E_{cum}$ of the winner neuron set $\mathrm{S_{winner}}$ in conv layers. 
\textit{Singular value decomposition} (SVD) is adopted to derive the eigenvalues for the feature map $fm$ and thereby seek for the optimal $p$. 
Before deriving SVD, 
$fm$ is reshaped into a matrix $M^{fm}\in \Re^{S\times C}$ by arranging all the activation vectors $fm[h,w,:]$ at each pixel position $[h,w]$ into rows, where $S$ is the total number of activation vectors. 
$S$ is then enlarged by gathering all the feature maps for a random set of training images (1,000 is enough by our experiments). 
Apply SVD on $M^{fm}$ such as
\begin{equation}
M_{S\times C}^{fm}=U_{S\times S}\Sigma_{S\times C} V_{C\times C}^{T}, 
\end{equation}
where $U$ and $V^{T}$ are unitary matrices, and $\Sigma$ is a diagonal matrix comprising the eigenvalues $\Lambda$. 
$M^{fm}$ can be converted into a new feature space as $M^{fm}V$, which is equal to $U_{S\times S}\Sigma_{S\times C}$. 
Due to the low-rank behavior of the responses of filters in the conv layer~\cite{zhang2016accelerating}, $\Sigma_{S\times C}$ can be reduced to $\Sigma_{S\times C^\prime}$, where $C^\prime$ is much smaller than $C$ by removing negligible eigenvalues. 
The winner rate $p$ can be determined by setting an appropriate cumulative energy threshold $\theta$ for the eigenvalues in $\Sigma$. Once $p$ is set, the winner set $\mathrm{S_{winner}}$ can be selected using the two-step ranking method aforementioned. 
After finetuning, the DASNet is trained to obtain the sparse feature representation $U_{S\times S}\Sigma_{S\times C^\prime}$. 

\begin{figure}[b]
\begin{minipage}{\columnwidth}
\centering
\vspace{-9pt}
\subfigure[The Lenet-4 for MNIST.]{
\includegraphics[width=0.7\columnwidth]{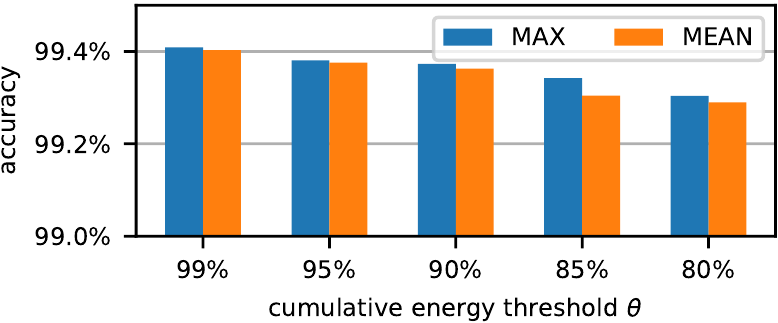}}
\subfigure[The ConvNet-5 for CIFAR-10.]{
\includegraphics[width=0.7\columnwidth]{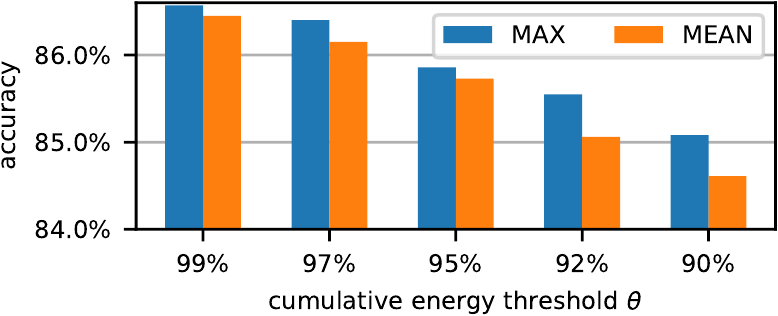}}
\vspace{-6pt}
\caption{Comparison between different feature vectors.}
\label{fig:max-vs-mean}
\end{minipage}
\end{figure}

We adopt LeNet-4 and ConvNet-5 in Table \ref{tab:toy_models} to evaluate the effectiveness of different feature vectors as defined in Equations (\ref{eq:mean}) and (\ref{eq:max}). 
For the comparison between two $fv$ extraction methods, the WTA masks are configured with identical winner rates according to the cumulative energy threshold $\theta$ settings. 
Interestingly, the $\mathbf{max}(\cdot)$ method outperforms the $\mathbf{mean}(\cdot)$ method regarding to the model accuracy as shown in Fig. \ref{fig:max-vs-mean}. 
As $\theta$ is approaching 100\%, the accuracy gap between the two methods decreases. 
In the following experiments, $\mathbf{max}(\cdot)$ will be adopted to derive the feature vector $fv$ if not specially indicated. 

\begin{figure}[t]
\centering
\includegraphics[width=0.8\columnwidth]{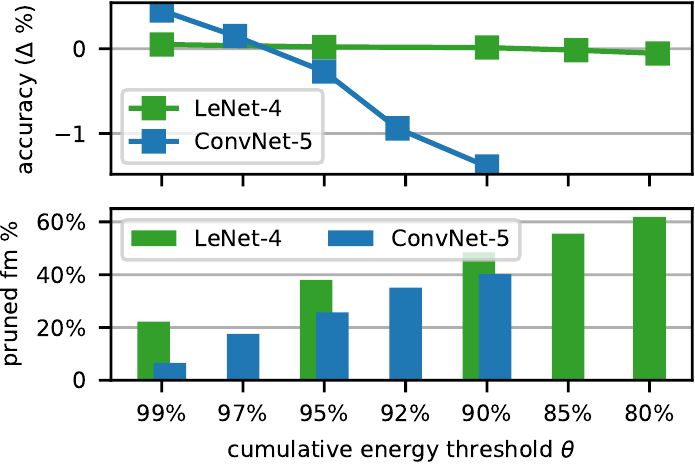}
\vspace{-6pt}
\caption{Accuracy vs. feature map pruning for conv layers.}
 \vspace{-10pt}
\label{fig:acc_fm}
\end{figure}

Fig. \ref{fig:acc_fm} shows that different cumulative energy threshold $\theta$ for $\mathrm{S_{winner}}$ results in different sparsity levels in the feature map. 
Lowering $\theta$ can aggressively increase the feature map sparsity. 
Again, we adopt different $\theta$ for LeNet-4 and ConvNet-5 for their different pruning sensitivity. 
Similar to the findings in fc layers, the accuracy drops along with the increment of sparsity level. 
As can be seen in Fig. \ref{fig:acc_fm}, 61.9\% of the feature maps among all conv layers can be pruned for LeNet-4 with a mere 0.05\% accuracy drop, while ConvNet-5 can prune 35.1\% of the feature maps by keeping the accuracy loss by 1\%. 
The feature map has a less pruning strength than the activations in fc layers because the feature map is more complicated in the feature space. 

\subsection{Theoretical Analysis}
\label{sec:analysis}


\subsubsection{Ranking cost} 
\label{sec:ranking_cost}
A lightweight ranking process is necessary to guarantee the benefit of saved computation via activation sparsification. 
The ranking overhead should be much less compared to the saved matrix computational complexity.
Assume a fc layer has $K$ inputs and $O$ outputs as illustrated in Fig.~\ref{fig:fc_wta}(a). 
Its computational complexity is $KO$.
After applying the dynamic WTA dropout, only the winner neurons from the previous layer, $pK$ input activations, are kept. 
So the lightened computation complexity is $pKO$. 
The ratio of the ranking cost over the saved computation complexity of the fc layer is:
\begin{equation}
\small
\begin{split}
\Big(\frac{Ranking~Cost}{Saved~Comp.}\Big)_{fc} & =\frac{K\log K}{(1-p)KO} \\
 & =\frac{\log K}{(1-p)O} \ll 1, 
\end{split}
\end{equation}
as $K$ and $O$ are usually in the similar magnitudes and $p$ in fc layers can be smaller than 20\% according to our experiments. 

Assume a conv layer has $N$ filters with the same configuration as shown in Fig. \ref{fig:conv_wta}. 
Its computation complexity originally is $F^{2}HWCN$ and reduces to $pF^{2}HWCN$ after applying the WTA dropout with a winner rate of $p$. 
Here the ranking cost is composed of two components: finding the maximum in each feature map channel with a complexity of $HWC$ and ranking with a complexity of $C\log C$. 
So the ratio of the ranking cost to the saved computational complexity for the conv layer is: 
\begin{equation}
\small
\begin{split}
\Big(\frac{Ranking~Cost}{Saved~Comp.}\Big)_{conv} & 
=\frac{HWC+C\log C}{(1-p)F^2HWCN} \\
    & \approx \frac{HW}{(1-p)F^2HWN} \\
    & =\frac{1}{(1-p)F^2N}\ll1 ,
\end{split}
\end{equation}
\noindent where $p$ of conv layers commonly is less than 70\% according to our experiments.

\subsubsection{Memory accesses and computational complexity}

The computations in DNNs are intensive matrix multiplications. 
From a mathematical view, the matrix computation in layer $l$ works as $W_l\cdot X_l$, 
where $W_l$ denotes the weight matrix, $X_l$ is the input vector/matrix derived from the previous layer's activations. 
The dynamic WTA method attempts to improve the sparsity level of the neuron activation. As such, the number of MACs involved is reduced. 
Moreover, sparse $X_l$ can reduce the memory allocation. 

For fc layers, it's straightforward to reduce the computation cost by $1/p$ times with a winner rate $p$ in vector $X_l$. 
For conv layers, the gains in the reduction of memory accesses and MACs can also be obtained from the WTA dropout. 
As the example depicted in Fig. \ref{fig:conv_wta}(a), the dimension of filters is $F\times F$, where $F$ is the window size of the 2-D convolution sliding over the feature map. 
When the convolution stride is 1, the feature map will be unrolled into a matrix $X_l$ in the dimension of $F^2C \times HW$. 
After applying the dynamic WTA dropout, the feature map will contain structured sparse patterns, where the matrix $X_l$ derived from the feature map unrolling can be condensed in the row direction. 
With a winner ratio of $p$ in the conv layer, both the data volume of $X_l$ and the MACs needed in $W_l\cdot X_l$ reduce $1/p$ times. 
Very importantly, the method doesn't require any specific compression techniques, e.g., compressed sparse row. 

\subsubsection{WTA mask reuse in backpropagation}

\begin{figure}[t]
\centering
\includegraphics[width=0.9\columnwidth]{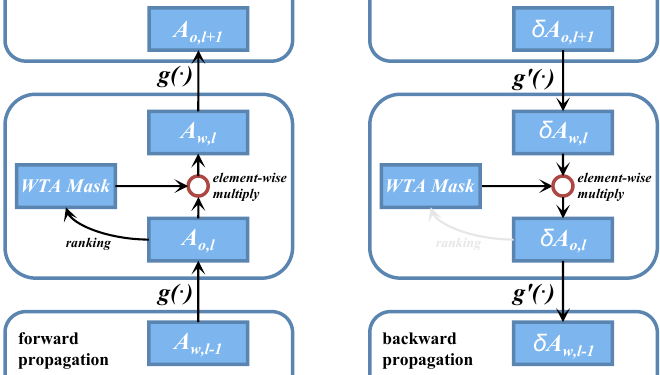}
\vspace{-3pt}
\caption{WTA mask in the forward and backward propagation.}
\vspace{-9pt}
\label{fig:mask-reuse}
\end{figure}

The reduction in memory accesses and computational complexity is also applicable to the backpropagation phase. 
As shown in the forward propagation graph in Fig. \ref{fig:mask-reuse}, the WTA mask is generated at run-time by ranking the activations $A_{o,l}=g(A_{w,l-1})$, where the function of layer $l$ is modeled as $g(\cdot)$. 
Only the activations from winner neurons are propagated to the next layer $l+1$, i.e., $A_{o,l}$ is element-wise multiplied with the WTA mask to obtain $A_{w,l}$. 
In the training process, the errors $\delta A_{w,l}=\frac{\partial Loss}{\partial A_{w,l}}$ and $\delta A_{o,l}=\delta A_{w,l}\cdot \frac{\partial A_{w,l}}{\partial A_{o,l}}$, where $\frac{\partial A_{w,l}}{\partial A_{o,l}}$ is the WTA mask, are needed to be backward propagated to derive weight updates as shown in Algorithm \ref{alg1}. 
Consistent with the dimension reduction in $A_{w,l}$ by the WTA dropout, the size of $\delta A_{w,l}$ is reduced accordingly. 
The WTA mask is reused as shown in the backward propagation graph in Fig. \ref{fig:mask-reuse}.
So there is no ranking overhead in the back-propagation phase. 

\subsection{DASNet Finetuning Flow}

\begin{figure}[t]
\centering
\includegraphics[width=0.85\columnwidth]{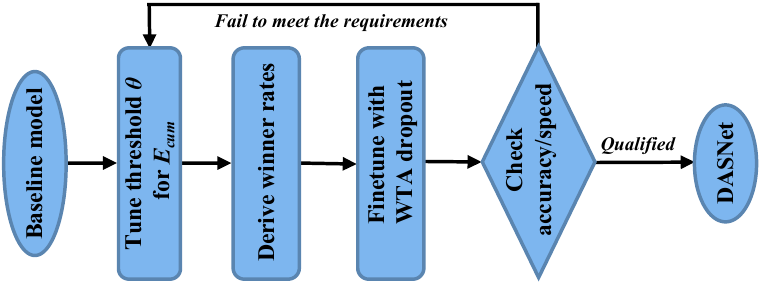}
\vspace{-3pt}
\caption{The finetuning flow to obtain DASNet.}
\vspace{-9pt}
\label{fig:tune-process}
\end{figure}

In practice, a finetuning flow in Fig. \ref{fig:tune-process} is used to obtain the final DASNet from the pretrained baseline model. 
Section \ref{sec:Approach} shows the tradeoff between the activation pruning strength and model accuracy. 
To meet the various requirements on prediction accuracy and inference speed for embedded applications, the finetuning procedure is iterated several times through tuning the cumulative energy threshold $\theta$. 
In all the experiments as we shall show in the following section, the accuracy drop for all the DASNets is kept within $0.5\%$. 
These models can be further accelerated by releasing the constraint on $\theta$ if the acceptable accuracy loss is larger. 
While more MAC reduction can be achieved with a smaller $\theta$, the DASNets are more vulnerable to accuracy loss. 

%% file: evaluation.tex
\section{Evaluation}
\label{sec:Evaluation}

\subsection{Model Setup}

\begin{table*}[t]
\caption{The experiment setup and results.}
\label{tab:experiments}
\centering
\begin{tabular}{c|c|ccc|ccc|c}
\hline\hline
Model                      & Dataset  & MAC \# & Top-1 Accuracy & Top-5 Accuracy & MAC \# Reduction & $\Delta$Top-1 Accuracy & $\Delta$Top-5 Accuracy & Speedup      \\
ResNet-164                 & CIFAR-10 & 384.5M & 94.54\%        & -              & 32.2\%           & -0.14\%                & -                      & 1.5$\times$ \\
\hline
\multirow{2}{*}{AlexNet}   & GTSRB    & 724.3M & 97.21\%        & -              & 34.9\%           & -0.5\%                 & -                      & 1.8$\times$ \\
& ImageNet & 748.1M & 57.22\%        & 80.2\%         & 27\%             & +0.41\%                & +0.12\%                & 1.6$\times$  \\
\hline
\multirow{2}{*}{MobileNet} & SVHN     &    11.9M    &        93.3\%        & -              &         45\%     &     -0.1\%           &             -           &    1.75$\times$ \\
   & ImageNet & 610.4M & 70.9\%         & 89.9\%         & 12\%         & -0.39\%                & +0\%                   & 1.12$\times$ \\
\hline\hline
\end{tabular}
\end{table*}

We verify the WTA dropout on real-world DNN models in Table \ref{tab:experiments}: ResNet \cite{he2016deep}, AlexNet and MobileNet~\cite{howard2017mobilenets}. 
Compared to conventional CNN models like ResNet and AlexNet, MobileNet is designed specially for embedded systems featuring a conv block by concatenating a depth-wise conv layer and a point-wise conv layer. 
MobileNet consists of 1 conv layer, 13 optimized conv blocks and 1 fc layer. 
We adopt four datasets including CIFAR-10, GTSRB (\textit{German Traffic Sign Recognition Benchmark}), SVHN (\textit{Street View House Number}) and ImageNet.
Besides the CIFAR-10,  
GTSRB has 43 classes of traffic signs, SVHN is a color-version of MNIST with 10 digits collected from natural scene images, and ImageNet contains 1000 class labels. 
The number of MACs is measured under minibatch size = 1, which is the typical scenario for real-time applications in embedded systems. 

\subsection{DASNet Measurement Results}

Our DASNets are deployed and experimented on LG Nexus 5X with a 1.8 GHz processor and 2GB RAM, running Android 6.0.1 (API level 23). 
We deploy the neural network models based on MXNet \cite{chen2015mxnet}, a high performance deep learning library developed by \textit{Distributed Machine Learning Community} (DMLC) team using C++. 
We modified the MXNet library and cross compiled it with Android Standalone Toolchain so that it can support \textit{general matrix multiplication} (GEMM) operations with our customized WTA dropout and neuron activation sparsity features on ARM-based platforms. 

As summarized in Table \ref{tab:experiments}, $1.12\times\sim 1.8\times$ wall-clock time speedup is achieved, which indicates that the structured sparsity in DASNets is easily to be utilized for acceleration. 
The DASNet acts differently on different datasets. 
For example, The speedup for AlexNet on GTSRB is $1.8\times$ when keeping $65.1\%$ MACs and the accuracy drops $0.5\%$. 
When changing the dataset to ImageNet, AlexNet even has a $0.41\%$ ($0.12\%$) improvement for Top-1 (Top-5) accuracy with a $1.6\times$ speedup. 
The MobileNet is more sensitive to the WTA dropout method. 
Through the SVD analysis on the feature maps in MobileNet on ImageNet, less redundancy is observed as in AlexNet. 
By saving $12\%$ MACs according to $\theta=99\%$, the inference is accelerated by $1.12\times$ with an $0.39\%$ drop for Top-1 accuracy, and it has no loss for the Top-5 accuracy. 
In contrast, more feature maps can be pruned for MobileNet on the less complex dataset SVHN, and thus a higher speedup ($1.75\times$) is achieved. 

\subsection{Case Study on Winner Rate Configuration}

\begin{figure}[t]
\begin{minipage}{\columnwidth}
\centering
\subfigure[The conv layers.]{
\includegraphics[width=0.9\columnwidth]{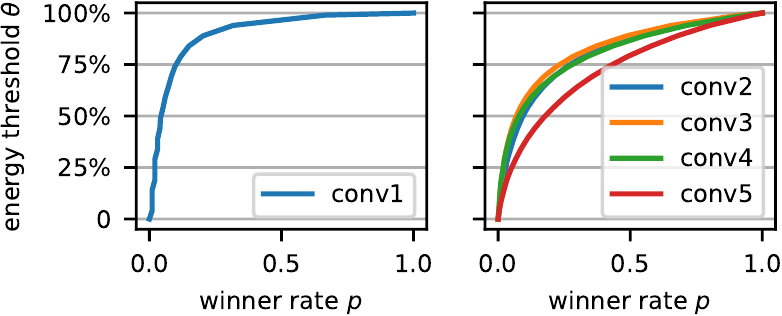}}
\subfigure[The fc layers.]{
\includegraphics[width=0.9\columnwidth]{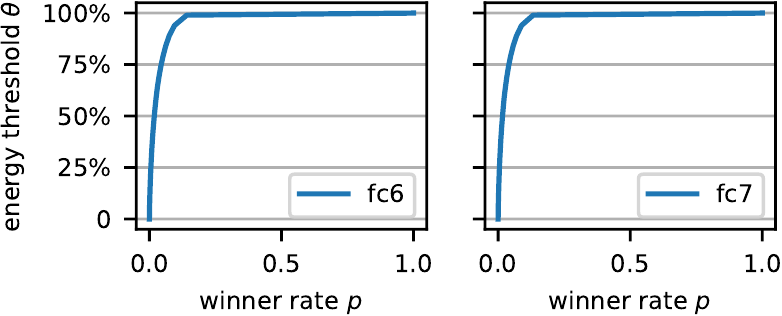}}
\vspace{-3pt}
\caption{An example of winner rate analysis for AlexNet.}
\label{fig:wr_analysis}
\end{minipage}
\vspace{-9pt}
\end{figure}

Before setting winner rate $p$ per layer, the relation between $p$ and the threshold $\theta$ of $E_{cum}$ is analyzed for each network. 
Similar to the setup in Section~\ref{sec:winner_select}, 1,000 images are randomly selected from the training set to analyze the selection of $p$ for feature maps in conv layers and activations in fc layers. 
Without loss of generality, the relation of $p$ and $\theta$ for AlexNet on ImageNet dataset is shown as an example in Fig.~\ref{fig:wr_analysis}. 
Since layer fc8 is the output layer, the analysis on the activations in fc layers are only applied for fc6 and fc7.

\begin{figure}[t]
\begin{minipage}{\columnwidth}
\centering
\subfigure[AlexNet on ImageNet.]{
\includegraphics[width=0.85\columnwidth]{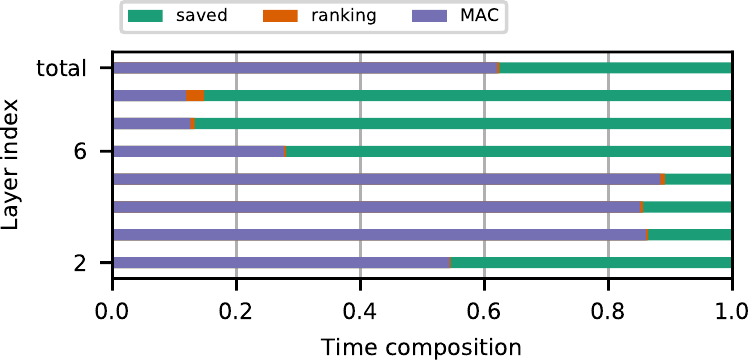}}
\subfigure[MobileNet on ImageNet.]{
\includegraphics[width=0.85\columnwidth]{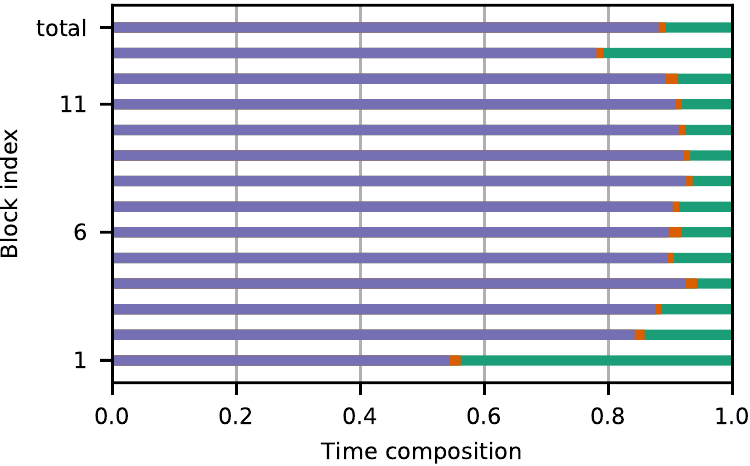}}
\vspace{-3pt}
\caption{The layer-wise decomposition for the inference time in DASNets. The input layer is omitted here, because the WTA dropout is not applicable on the input data.}
\label{fig:per_layer}
\end{minipage}
\vspace{-9pt}
\end{figure}

As seen from the analysis results, different layers have distinct behavior of winner rates. 
For example, the winner rate $p$ of conv1 is $0.54$ by setting $\theta = 99\%$, while $p=0.72$ for conv2 at the same $\theta$. 
The divergence in the derived $p$ values indicates the different feature redundancy in the generated feature space per layer. 
It's worth noting that the cumulative energy in fc layers increases rapidly with the increment of winner rate. 
When $p$ is set around $0.1$, $E_{cum}$ can be easily kept more than $95\%$, which means the activations of fc layers can be deeply sparsified. 
The analysis in Fig. \ref{fig:wr_analysis} can be generalized to different combinations of datasets and model structures to explore activation sparsity. 
Thereafter, the DASNet with WTA dropout will efficiently utilize the improved activation sparsity derived from the winner rate analysis.

\subsection{Case Study on Layer-wise Speedup}

The speedup results for each layer in AlexNet and MobileNet on ImageNet are shown in Fig. \ref{fig:per_layer}. 
The time consumption per layer consists of both the ranking process and the remaining MAC computation for the winner neurons. 
For AlexNet and MobileNet, no WTA dropout is applied on the input image. 
To have a clear view of time distribution among all layers, the time is normalized to the computation time of the original dense layer. 
As shown in Fig.~\ref{fig:per_layer}(a), for conv layers 2-5 in AlexNet, $1.13\times\sim 1.84\times$ speedup is achieved. It's worth noting that fc layers 6-8 obtain better speedup results of $3.6\times\sim7.6\times$. This is due to that the WTA masks before fc layers usually generate a deeply sparsified input. 
For MobileNet, the 13 conv blocks account for more than $90\%$ computation of the whole network. 
As the results shown in Fig.~\ref{fig:per_layer}(b), $1.07\times\sim 1.96\times$ speedup per block is obtained. 
For both AlexNet and MobileNet, the time consumption on ranking process can be negligible,  which accounts for a small portion ($<2\%$) of the original model execution time.

\subsection{Integration with Weight Pruning or Quantization}

A significant feature of DASNets is that they can be seamlessly integrated with weight pruning and quantization. 
To explore the integration with weight pruning, the three dense fc layers in AlexNet are adopted as shown in Table \ref{tab:w_prune}, which account for $\sim96\%$ of the total model size. 
After successfully compressing the fc weights by $5.5\times$ without accuracy loss, two cases of cumulative energy threshold $\theta$ are applied for the WTA dropout. 
When $\theta=99\%$, $79.6\%$ of the input activations can be masked out with no accuracy loss. 
To further improve the activation sparsity using $\theta=95\%$, only $18.1\%$ activations are kept with a mere $0.07\%$ Top-1 accuracy drop. 
To explore the integration with quantization, the 8-bit linear quantization as \cite{jouppi2017datacenter} is adopted. All quantized models are derived at one-shot without intricate finetuning. 
For the AlexNet on GTSRB in Table \ref{tab:experiments}, the corresponding 8-bit DASNet has little accuracy loss ($<0.01\%$) compared to the 32-bit floating-point counterpart. 
For the ImageNet dataset, the 8-bit DASNets for AlexNet and MobileNet have marginal $0.5\%$ and $0.8\%$ accuracy loss respectively. 

\begin{table}[t]
\caption{Integration with weight pruning.}
\label{tab:w_prune}
\vspace{-3pt}
\centering
\resizebox{1\columnwidth}{!}
{\begin{minipage}{\columnwidth}
\centering
\begin{tabular}{c|cc|cc}
\hline\hline
\multirow{2}{*}{Layer} & \multirow{2}{*}{Weights} & \multirow{2}{*}{Weights \%} & \multicolumn{2}{c}{Actications \%} \\ \cline{4-5} 
&   &  & $\theta$=99\%    & $\theta$=95\%   \\
\hline
fc6  & 37.75M     & 17.2\%     & 27.7\%  & 25.6\%  \\
fc7   & 16.78M & 17.2\%  & 12.5\% & 10.1\%          \\
fc8  & 4.10M     & 31\%    & 11.9\%   & 9.4\%  \\
\hline
Total   & 58.63M    & 18.2\%      & 20.4\%  & 18.1\% \\
\hline\hline
\end{tabular}
\end{minipage}}
\vspace{-10pt}
\end{table}

\subsection{Comparison with State-of-the-art}

Many methods have been proposed for feature map pruning. 
The comparison between the our WTA dropout method with state-of-the-art is shown in Table \ref{tab:comparison}. 
The models denoted with `$\checkmark$' dynamically prune feature maps in contrast to the static pruning in others. 
Compared to the static pruning methods, our WTA dropout method obtains a better accuracy with similar reductions of MACs. 
In the cases with a small computation save, DASNet models even boost the accuracy over the original models. 
For the existing dynamic feature map pruning methods, they actually complicate network structures and lack experiments on deeper models than ResNet-18. 
Our WTA dropout method is more supportive to DNNs as it doesn't increase training variables. 
Compared with the existing dynamically pruned ResNet-18, the accuracy drop of DASNet is smaller with comparable computation reduction. 

\begin{table}[t]
\centering
\caption{Comparison with existing feature map pruning methods.}
\label{tab:comparison}
\scriptsize
\resizebox{1.0\columnwidth}{!}
{\begin{minipage}{\columnwidth}
\centering
\begin{tabular}{cclccc}
\hline\hline
Model       &  \begin{tabular}{@{}c@{}} Top5 \\ Accuracy \end{tabular}          & Pruned Model    & \begin{tabular}{@{}c@{}} $\Delta$ Top5 \\ Accuracy \end{tabular}  & \begin{tabular}{@{}c@{}} MAC \# \\ Reduction \end{tabular} \\
\hline
\multirow{2}{*}{AlexNet}   & \multirow{2}{*}{80.2\%} & Ours ($\checkmark$)  & +0.12\%          & 27\%             \\
    &       & Molchanov et al.  & -1.4\%        & 22.4\%           \\
\hline
\multirow{3}{*}{ResNet-18} & \multirow{3}{*}{89.68\%} &     Hua et al. ($\checkmark$) & -1.87\% &  37.9\%   \\
    &  &        Gao et al. ($\checkmark$)     & -1.46\%    & 49.5\% \\
    &  &        Ours ($\checkmark$) & -1.24\% &  50.4\% \\
\hline
\multirow{3}{*}{ResNet-50} & \multirow{3}{*}{92.8\%} 
   &    Ours-v1 ($\checkmark$)               & +0.26\%                       &     15.9\%           \\
   &       & Huang \& Wang   & -0.19\%                  & 15.1\%           \\
   &      & Ours-v2 ($\checkmark$)      &     -0.37\%        & 29\%             \\
\hline\hline
\end{tabular}
\end{minipage}}
\vspace{-9pt}
\end{table}

%% file: conclusion.tex
\section{Conclusion}
\label{sec:Conclusion}

In this paper, we propose the dynamic WTA dropout to generate structured sparsity in feature map for conv layers and deeply sparsify the activations for fc layers. 
The DASNet equipped with dynamic WTA dropout can be efficiently utilized by conventional embedded systems without requiring dedicated hardware. 
The proposed WTA dropout is a generic approach to explore the dynamic activation sparsity which can be easily applied to DNN applications beyond the models and datasets in this paper. 
Our experiments on the mobile platform show an $1.12\times\sim1.8\times$ speedup on various DNN models by keeping a $0.5\%$ accuracy loss. 
The DASNets can also be integrated with weight pruning and quantization without compromising on accuracy, which presents a great potential for the efficient DNN deployment in embedded systems.

%% file: main.bbl
\begin{thebibliography}{10}
\providecommand{\url}[1]{#1}
\csname url@samestyle\endcsname
\providecommand{\newblock}{\relax}
\providecommand{\bibinfo}[2]{#2}
\providecommand{\BIBentrySTDinterwordspacing}{\spaceskip=0pt\relax}
\providecommand{\BIBentryALTinterwordstretchfactor}{4}
\providecommand{\BIBentryALTinterwordspacing}{\spaceskip=\fontdimen2\font plus
\BIBentryALTinterwordstretchfactor\fontdimen3\font minus
  \fontdimen4\font\relax}
\providecommand{\BIBforeignlanguage}[2]{{%
\expandafter\ifx\csname l@#1\endcsname\relax
\typeout{** WARNING: IEEEtran.bst: No hyphenation pattern has been}%
\typeout{** loaded for the language `#1'. Using the pattern for}%
\typeout{** the default language instead.}%
\else
\language=\csname l@#1\endcsname
\fi
#2}}
\providecommand{\BIBdecl}{\relax}
\BIBdecl

\bibitem{krizhevsky2012imagenet}
A.~Krizhevsky, I.~Sutskever, and G.~E. Hinton, ``Imagenet classification with
  deep convolutional neural networks,'' in \emph{Advances in neural information
  processing systems}, 2012, pp. 1097--1105.

\bibitem{jouppi2017datacenter}
N.~P. Jouppi, C.~Young, N.~Patil, D.~Patterson, G.~Agrawal, R.~Bajwa, S.~Bates,
  S.~Bhatia, N.~Boden, A.~Borchers \emph{et~al.}, ``In-datacenter performance
  analysis of a tensor processing unit,'' in \emph{Proceedings of the ACM/IEEE
  International Symposium on Computer Architecture}, 2017, pp. 1--12.

\bibitem{courbariaux2015binaryconnect}
M.~Courbariaux, Y.~Bengio, and J.-P. David, ``Binaryconnect: Training deep
  neural networks with binary weights during propagations,'' in \emph{Advances
  in neural information processing systems}, 2015, pp. 3123--3131.

\bibitem{hubara2016binarized}
I.~Hubara, M.~Courbariaux, D.~Soudry, R.~El-Yaniv, and Y.~Bengio, ``Binarized
  neural networks,'' in \emph{Advances in neural information processing
  systems}, 2016, pp. 4107--4115.

\bibitem{rastegari2016xnor}
M.~Rastegari, V.~Ordonez, J.~Redmon, and A.~Farhadi, ``Xnor-net: Imagenet
  classification using binary convolutional neural networks,'' in
  \emph{Proceedings of the European Conference on Computer Vision}, 2016, pp.
  525--542.

\bibitem{horowitz20141}
M.~Horowitz, ``Computing's energy problem (and what we can do about it),'' in
  \emph{Proceedings of the IEEE International Solid-State Circuits Conference},
  2014, pp. 10--14.

\bibitem{wen2016learning}
W.~Wen, C.~Wu, Y.~Wang, Y.~Chen, and H.~Li, ``Learning structured sparsity in
  deep neural networks,'' in \emph{Advances in Neural Information Processing
  Systems}, 2016, pp. 2074--2082.

\bibitem{molchanov2017pruning}
P.~Molchanov, S.~Tyree, T.~Karras, T.~Aila, and J.~Kautz, ``Pruning
  convolutional neural networks for resource efficient inference,'' in
  \emph{Proceedings of the International Conference on Learning
  Representations}, 2017.

\bibitem{han2015deep}
S.~Han, H.~Mao, and W.~J. Dally, ``Deep compression: Compressing deep neural
  networks with pruning, trained quantization and huffman coding,'' \emph{arXiv
  preprint arXiv:1510.00149}, 2015.

\bibitem{chen2017eyeriss}
Y.-H. Chen, T.~Krishna, J.~S. Emer, and V.~Sze, ``Eyeriss: An energy-efficient
  reconfigurable accelerator for deep convolutional neural networks,''
  \emph{IEEE Journal of Solid-State Circuits}, vol.~52, no.~1, pp. 127--138,
  2017.

\bibitem{rhu2018compressing}
M.~Rhu, M.~O'Connor, N.~Chatterjee, J.~Pool, Y.~Kwon, and S.~W. Keckler,
  ``Compressing dma engine: Leveraging activation sparsity for training deep
  neural networks,'' in \emph{Proceedings of the IEEE International Symposium
  on High Performance Computer Architecture}, 2018, pp. 78--91.

\bibitem{han2016eie}
S.~Han, X.~Liu, H.~Mao, J.~Pu, A.~Pedram, M.~A. Horowitz, and W.~J. Dally,
  ``Eie: efficient inference engine on compressed deep neural network,'' in
  \emph{Proceedings of the ACM/IEEE International Symposium on Computer
  Architecture}, 2016, pp. 243--254.

\bibitem{kim2018zena}
D.~Kim, J.~Ahn, and S.~Yoo, ``Zena: Zero-aware neural network accelerator,''
  \emph{IEEE Design \& Test}, vol.~35, no.~1, pp. 39--46, 2018.

\bibitem{albericio2016cnvlutin}
J.~Albericio, P.~Judd, T.~Hetherington, T.~Aamodt, N.~E. Jerger, and
  A.~Moshovos, ``Cnvlutin: Ineffectual-neuron-free deep neural network
  computing,'' in \emph{Proceedings of the ACM/IEEE International Symposium on
  Computer Architecture}, 2016, pp. 1--13.

\bibitem{reagen2016minerva}
B.~Reagen, P.~Whatmough, R.~Adolf, S.~Rama, H.~Lee, S.~K. Lee, J.~M.
  Hern{\'a}ndez-Lobato, G.-Y. Wei, and D.~Brooks, ``Minerva: Enabling
  low-power, highly-accurate deep neural network accelerators,'' in
  \emph{Proceedings of the ACM/IEEE International Symposium on Computer
  Architecture}, 2016, pp. 267--278.

\bibitem{spring2017scalable}
R.~Spring and A.~Shrivastava, ``Scalable and sustainable deep learning via
  randomized hashing,'' in \emph{Proceedings of the ACM International
  Conference on Knowledge Discovery and Data Mining}, 2017, pp. 445--454.

\bibitem{liu2017learning}
Z.~Liu, J.~Li, Z.~Shen, G.~Huang, S.~Yan, and C.~Zhang, ``Learning efficient
  convolutional networks through network slimming,'' in \emph{Proceedings of
  the IEEE International Conference on Computer Vision}, 2017, pp. 2755--2763.

\bibitem{ye2018rethinking}
J.~Ye, X.~Lu, Z.~Lin, and J.~Z. Wang, ``Rethinking the
  smaller-norm-less-informative assumption in channel pruning of convolution
  layers,'' \emph{arXiv preprint arXiv:1802.00124}, 2018.

\bibitem{huang2018data}
Z.~Huang and N.~Wang, ``Data-driven sparse structure selection for deep neural
  networks,'' in \emph{Proceedings of the European Conference on Computer
  Vision}, 2018, pp. 304--320.

\bibitem{hua2018channel}
W.~Hua, C.~D. Sa, Z.~Zhang, and G.~E. Suh, ``Channel gating neural networks,''
  \emph{http://arxiv.org/abs/1805.12549}, 2018.

\bibitem{gao2018dynamic}
X.~Gao, Y.~Zhao, Łukasz Dudziak, R.~Mullins, and C.~zhong Xu, ``Dynamic
  channel pruning: Feature boosting and suppression,'' in \emph{International
  Conference on Learning Representations}, 2019.

\bibitem{bentley1980general}
J.~L. Bentley, D.~Haken, and J.~B. Saxe, ``A general method for solving
  divide-and-conquer recurrences,'' \emph{ACM SIGACT News}, pp. 36--44, 1980.

\bibitem{srivastava2014dropout}
N.~Srivastava, G.~Hinton, A.~Krizhevsky, I.~Sutskever, and R.~Salakhutdinov,
  ``Dropout: a simple way to prevent neural networks from overfitting,''
  \emph{The Journal of Machine Learning Research}, vol.~15, no.~1, pp.
  1929--1958, 2014.

\bibitem{zeiler2014visualizing}
M.~D. Zeiler and R.~Fergus, ``Visualizing and understanding convolutional
  networks,'' in \emph{Proceedings of the European conference on computer
  vision}, 2014, pp. 818--833.

\bibitem{bau2017network}
D.~Bau, B.~Zhou, A.~Khosla, A.~Oliva, and A.~Torralba, ``Network dissection:
  Quantifying interpretability of deep visual representations,'' \emph{arXiv
  preprint arXiv:1704.05796}, 2017.

\bibitem{zhang2016accelerating}
X.~Zhang, J.~Zou, K.~He, and J.~Sun, ``Accelerating very deep convolutional
  networks for classification and detection,'' \emph{IEEE transactions on
  pattern analysis and machine intelligence}, vol.~38, no.~10, pp. 1943--1955,
  2016.

\bibitem{he2016deep}
K.~He, X.~Zhang, S.~Ren, and J.~Sun, ``Deep residual learning for image
  recognition,'' in \emph{Proceedings of the IEEE conference on computer vision
  and pattern recognition}, 2016, pp. 770--778.

\bibitem{howard2017mobilenets}
A.~G. Howard, M.~Zhu, B.~Chen, D.~Kalenichenko, W.~Wang, T.~Weyand,
  M.~Andreetto, and H.~Adam, ``Mobilenets: Efficient convolutional neural
  networks for mobile vision applications,'' \emph{arXiv preprint
  arXiv:1704.04861}, 2017.

\bibitem{chen2015mxnet}
T.~Chen, M.~Li, Y.~Li, M.~Lin, N.~Wang, M.~Wang, T.~Xiao, B.~Xu, C.~Zhang, and
  Z.~Zhang, ``Mxnet: A flexible and efficient machine learning library for
  heterogeneous distributed systems,'' \emph{arXiv preprint arXiv:1512.01274},
  2015.

\end{thebibliography}
